\documentclass[9pt,twocolumn,twoside]{pnas-new}

 \templatetype{pnasresearcharticle} 

\title{Universal Multilayer Network Exploration by Random Walk with Restart}

\author[1,2,*]{Anthony Baptista}
\author[2]{Aitor Gonzalez}
\author[1,3,*]{Anaïs Baudot}

\affil[1]{Aix-Marseille Univ, INSERM, MMG, Turing Center for Living Systems, CNRS, Marseille, France}
\affil[2]{Aix-Marseille Univ, INSERM, TAGC, Turing Center for Living Systems, Marseille, France}
\affil[3]{Barcelona Supercomputing Center, Barcelona, Spain}

\correspondingauthor{* anthony.baptista@univ-amu.fr, anais.baudot@univ-amu.fr}
\leadauthor{A.Baptista}


\begin{abstract}
\section*{Abstract}
The amount and variety of data have been increasing drastically for several years. These data are often represented as networks and explored with approaches arising from network theory. Recent years have witnessed the extension of network exploration approaches to capitalize on more complex and richer network frameworks. Random walks, for instance, have been extended to explore multilayer networks. However, current random walk approaches are limited in the combination and heterogeneity of networks they can handle. New analytical and numerical random walk methods are needed to cope with the increasing diversity and complexity of multilayer networks. We propose here MultiXrank, a method and associated Python package that enables Random Walk with Restart on any kind of multilayer network. We evaluate MultiXrank with leave-one-out cross-validation and link prediction, and measure the impact of the addition or removal of network data on prediction performances. Finally, we measure the sensitivity of MultiXrank to input parameters by in-depth exploration of the parameter space.
\end{abstract}

\begin{document}

\maketitle
\thispagestyle{firststyle}
\ifthenelse{\boolean{shortarticle}}{\ifthenelse{\boolean{singlecolumn}}{\abscontentformatted}{\abscontent}}{}

\section*{Introduction}

\dropcap{D}ata amount and variety have soared as never seen before, offering a unique opportunity to better understand complex systems. Among the different modes of representation of data, networks appear as particularly successful. Networks are indeed interesting to refine raw data and extract relevant features, patterns, and classes. They are exploited for years to study complex systems, and a wide and powerful range of tools from graph theory are available for their exploration.\\
However, the integrated exploration of large multidimensional datasets remains a major challenge in many scientific fields. 
For instance, a comprehensive understanding of biological systems would require the integrated analysis of dozens of different datasets produced at different molecular, cellular or tissular scales.
Recently, multilayer networks emerged as essential players in the analysis of such complex systems. Multilayer networks allow integrating more than one network in a unified formalism, in which the different networks are considered as layers \cite{Bianconi2018}. For instance, Duran-Frigola et al. \cite{DuranFrigola2020} combined 25 different networks of chemical compounds and their relationships, gathering relationships from chemical structures to clinical outcomes. This multilayer framework allows an integrated study of chemical compounds and their biological activities. Another example is given by the Hetionet project. The authors collected dozen of heterogeneous networks, i.e networks with various types of nodes such as genes, drugs or diseases, to prioritize drugs for repurposing \cite{Himmelstein2017}. \\
Several definitions of multilayer networks have been proposed, based on the (in)homogeneity of the layers and the properties of the connections between layers \citep{DeDomenico2013, Kivelae2014, Lee2020}. For instance, multiplex networks are multilayer networks composed of different layers containing the same nodes (called replica nodes) but different types of edges, and thereby different topologies. Heterogeneous networks link networks composed of different types of nodes thanks to bipartite interactions. Temporal networks follow the dynamic of a network over time: all the layers have the same nodes, but each layer represents the interaction state at a given time \citep{Holme2012}. We will here consider universal multilayer networks, which can be defined as multilayer networks composed of any number of multiplex (or monoplex) networks (with edges that can be directed and/or weighted), linked by bipartite networks (with edges that can be directed and/or weighted) (Fig. 1).
A wide range of methods have been developed in the recent years to analyze multilayer networks. For instance, different network metrics have been adapted to multilayer networks \citep{Battiston2014}, as well as various network clustering algorithms for community detection \citep{Mucha2010, Didier2015, Choobdar2019} or random walk for network exploration \citep{Li2010, DeDomenico2014, Cho2016, Valdeolivas2018}. \\
Random walks are iterative stochastic processes widely used to explore network topologies. They can be described as simulated particles that walk iteratively from one node to one of its neighbors with some probability \cite{Lovasz1993}.
The PageRank algorithm, for instance, is based on a random walk simulating the behavior of an internet user walking from one page to another thanks to hyper-links. The user can also restart the walk on any arbitrary page \cite{Brin1998}. In this particular random walk strategy, the restart prevents the random walker from being trapped in dead-ends \cite{Langville2006}. An interesting alternative strategy restricts the restart to specific node(s), called the seed(s) \cite{Pan2004}. In this strategy, named Random Walk with Restart (RWR) or Personalized PageRank, the random walk represents a measure of proximity from all the nodes in the network to the seed(s). RWR can also be described as a diffusion process, in which the objective is to determine the steady-state of an initial probability distribution \citep{Gomez2013}.\\ 
RWR are widely used to exploit large-scale networks. In computational biology, for instance, RWR strategies have been shown to significantly outperform methods based on local distance measures for the prioritization of gene-disease associations \cite{Koehler2008}. Importantly, different upgrades of the RWR approach have been implemented during the last decade, including its extension to i) heterogeneous networks \cite{Li2010}, ii) multiplex networks \cite{DeDomenico2014} and iii) multiplex-heterogeneous networks \cite{Valdeolivas2018}. In RWR, the degrees of freedom are summarized in the Transition rate matrix, and correspond to the available transitions between the different nodes of the graph. The extensions of RWR are challenging because the Transition rate matrices need to be normalized. To the best of our knowledge, this normalization is currently only solved for multilayer networks composed of two heterogeneous multiplex networks \cite{Valdeolivas2018, PioLopez2021} and the more universal case of $N$ multiplex networks remains unsolved. \\
We propose here MultiXrank, a framework composed of a method and a Python package to execute RWR on universal multilayer networks. We first introduce the mathematical bases of this RWR for universal multilayer networks, which correspond to a generalization of the approach from \cite{Li2010}. We evaluated MultiXrank with leave-one-out cross-validation and link prediction protocols. These evaluations reveal that more network data is not always better and highlight the critical influence of the bipartite networks. We finally present an in-depth exploration of the parameter space to measure the stability of the RWR output scores under variations of the input parameters. The MultiXrank Python package is freely available at \href{https://github.com/anthbapt/multixrank}{https://github.com/anthbapt/multixrank}, with an optimized implementation allowing its application to large multilayer networks.

\section*{Results}

\subsection*{Random Walk with Restart (RWR)}
\hfill\\

Let us consider an irreducible and aperiodic Markov chain, for instance a network composed of a giant component with undirected edges, $G = (V, E)$, where $V$ is the set of vertices and $E \subseteq (V$*$V)$ is the set of edges. In the case of irreducible and aperiodic Markov chains, a stationary probability ${\bf p}^{*}$ exists and satisfies the following properties:

\begin{align}
    \left\{
    \begin{array}{ll}
    {\bf p}^{*}(i) > 0 \; ; \; \; \; \forall i \in V \\
    \sum_{i \in V} {\bf p}^{*}(i) = 1 \\
    \end{array}
    \right.
\end{align}

We next introduce the probability defining the walk from one node to another. Let us define $x$, a particle that explores the network, $x_{t}$ its position at time $t$ and $x_{t+1}$ its position at time $t+1$. Considering two nodes $i$ and $j$: 

\begin{align}
    \mathbb{P}(x_{t+1} = j \; | \; x_{t} = i) = \left\{
    \begin{array}{ll}
    \frac{1}{d_{i}} & \mbox{if $(i, j) \in E$} \\
    0 & \mbox{Otherwise}\\
    \end{array}
    \right.
\end{align}

\noindent with $d_{i}$ being the degree of the node $i$. All the normalized possible transitions can be included in the Transition rate matrix. This Transition rate matrix, noted $M$, can be seen as the matrix of the degrees of freedom of the particle in the system. It is useful to note that the Transition rate matrix is equal to the column-normalized Adjacency matrix.
The distribution denoted by ${\bf p}_{t} = ({\bf p}_{t}(i))_{i \in V}$ describes the probability of being in the node $i$ at time $t$, and the stationary distribution ${\bf p}^{*}$ is obtained thanks to the homogeneous linear difference equation [3] \cite{Meyer2000, Langville2006}: 

\begin{align}
	{\bf p}_{t+1}^{T} = M{\bf p}_{t}^{T}
\end{align}

\noindent with ${\bf p}_{t}^{T}$ denoting the transpose of the vector ${\bf p}_{t}$. Moreover, we can introduce a non-homogeneous linear difference equation [4] \cite{Meyer2000} to take into account the restart on the seed(s). When the Transition rate matrix is a Stochastic matrix, the stationary distribution is reached \cite{Langville2006} (Supplementary Note 1.A.1 for elements of proof of convergence) and this distribution can be seen as a measure of proximity of all the network nodes with respect to the seed(s). 

\begin{align}
	{\bf p}_{t+1}^{T} = (1-r)M{\bf p}_{t}^{T} + r{\bf p}_{0}^{T}
\end{align}

The distribution ${\bf p}_{0}$ corresponds to the initial probability distribution, where only the seed(s) have non-zero values; $r$ represents the restart probability.\\

\subsection*{RWR on Multiplex networks}
\hfill\\

The RWR method has been extended to multiplex networks, i.e., multilayer networks with a one-to-one mapping between the (replica) nodes of the different layers (Fig. 1) \cite{DeDomenico2014, Cho2016, Bianconi2018}.
Multiplex networks can be represented by Supra-adjacency matrices, which correspond to a generalization of the standard Adjacency matrix. In the following, we will use several multiplex networks, indexed by $k$. We denoted by $\mathcal{A}_{k}$ the Supra-adjacency matrix of the multiplex network indexed by $k$. The Adjacency matrix of the layer $l$ of the multiplex network $k$ is denoted by $A_{k}^{[l]}$. The element of this adjacency matrix from node $i$ to node $j$ is defined as $(A_{k}^{[l]})_{i,j} \geq 0$. The dimension of the Supra-adjacency matrix $\mathcal{A}_{k}$ of the multiplex network $k$ is equal to $(L_{k}*n_{k})$*$(L_{k}*n_{k})$, with $n_{k}$ the number of nodes in each layer of the multiplex network $k$ and $L_{k}$ the number of layers in the multiplex network $k$. The Supra-adjacency matrix $\mathcal{A}_{k}$ is defined as follows:

\begin{align}
  (\mathcal{A}_{k})_{i_{l} , j_{m}} = \left\{
    \begin{array}{ll}
    (A_{k}^{[l]})_{i, j} & \mbox{if $l = m$} \\
    \hspace{0.2cm} \delta_{i, j} & \mbox{if $l \neq m$} \\
    \end{array}
    \right.
\end{align}

\noindent where $\delta$ defines the Kronecker delta (i.e., 1 if $i$ equal $j$ and 0 otherwise), and $l$ and $m$ represent the layers of the multiplex network $k$. We can also define a multiplex network as a set of nodes, $V_{\mathcal{A}_{k}}$ and a set of edges, $E_{\mathcal{A}_{k}}$: 

\begin{align}
    \left\{
    \begin{array}{ll}
    	G_{\mathcal{A}_{k}} = (V_{\mathcal{A}_{k}}, E_{\mathcal{A}_{k}})\\
    	V_{\mathcal{A}_{k}} = \{v_{i}^{l}, i = 1, ..., n_{k}, l = 1, ..., L_{k}\}\\
    	E_{\mathcal{A}_{k}} = \{e_{i, j}^{l l}, i,j = 1, ..., n_{k}, l = 1, ..., L_{k},(A_{k}^{[l]})_{i, j} \neq 0\} \\
    	\hspace{1.2cm} \cup \{e_{i, i}^{l m}, i = 1, ..., n_{k}, l \neq m \} \\
    \end{array}
    \right.
\end{align}

Importantly, we need to column-normalize the Supra-adjacency matrix defined in the equations [5-6] in order to converge to the steady-state, as defined in \cite{Valdeolivas2018}. This normalization requires including the parameters $\delta_{k}$ related to the jumps from one layer to another inside the matrix representation, as described in \cite{DeDomenico2014} (Fig. 2).
In the next section, we need to index by $k$ all the parameters that are dedicated to the multiplex network $k$. The Supra-adjacency matrix representing the multiplex network $k$ can be written as described in equation [7]. The matrix $I_{k}$ represents the Identity matrix of size $n_{k}$. 

\begin{align}
    \mathcal{A}_{k} = 
    \begin{bmatrix}
    (1-\delta_{k})A_{k}^{[1]} & \frac{\delta_{k}}{(L_{k}-1)}I_{k} & ... & \frac{\delta_{k}}{(L_{k}-1)}I_{k} \vspace{0.3cm}\\
    \frac{\delta_{k}}{(L_{k}-1)}I_{k} & (1-\delta_{k})A_{k}^{[2]} & ... & \frac{\delta_{k}}{(L_{k}-1)}I_{k} \vspace{0.3cm}\\
    ... & ... & ... & ... \vspace{0.3cm}\\
    \frac{\delta_{k}}{(L_{k}-1)}I_{k} & \frac{\delta_{k}}{(L_{k}-1)}I_{k} & ... & (1-\delta_{k})A_{k}^{[L_{k}]} \vspace{0.3cm}\\
    \end{bmatrix}
\end{align}

\subsection*{RWR on universal multilayer networks}
\hfill\\

We here define a RWR method that can be applied to universal multilayer networks. Universal multilayer networks are composed of any combination of multiplex networks, linked by any combination of bipartite networks (Fig. 1). All network edges can also be weighted and/or directed. The formalism for the application of RWR on multiplex networks is described in the previous section. We will now detail the Bipartite network matrices, and how to combine intra- and inter- multiplex networks information to obtain the Supra-heterogeneous adjacency matrix. The Supra-heterogeneous adjacency matrix will embed all the possible transitions in a universal multilayer network.

\subsubsection*{Bipartite networks connect heterogeneous nodes}
\hfill\\

The Bipartite network matrices contain the transitions between different types of nodes present in different networks. If the network $\alpha$ has $n_{\alpha}$ nodes, and the network $\beta$ has $n_{\beta}$ nodes, the Bipartite network matrix denoted $b_{\alpha,\beta}$ has a size equal to $n_{\alpha}*n_{\beta}$. Now, let us define $\mathcal{A}_{\alpha}$ and $\mathcal{A}_{\beta}$, two Supra-adjacency matrices representing the multiplex networks $\alpha$ and $\beta$. The Bipartite network matrix $B_{\alpha, \beta}$ represents the transitions from the nodes of the multiplex network $\alpha$ to the nodes of the multiplex network $\beta$. The size of the Bipartite network matrix $B_{\alpha, \beta}$ is equal to $(L_{\alpha}*n_{\alpha})$*$(L_{\beta}*n_{\beta})$. The Bipartite network matrices are composed of $(L_{\alpha}*L_{\beta})$ times the Bipartite network matrix $b_{\alpha,\beta}$ (equation [8]). The matrix $b_{\alpha,\beta}$ is composed of all the transitions from one layer of the multiplex network $\alpha$ to one layer of the multiplex network $\beta$.
We extended the formalism used in \cite{Valdeolivas2018} in order to consider more than two different multiplex networks.

\begin{align}
  B_{\alpha,\beta} = \left.
    \vphantom{\begin{array}{c}1\\1\\1\\1\\1\\1\\1\\1\end{array}}
    \smash{\underbrace{	
    		\begin{bmatrix}
    		b_{\alpha,\beta} & b_{\alpha,\beta} & ... & b_{\alpha,\beta} \vspace{0.3cm}\\
    		b_{\alpha,\beta} & b_{\alpha,\beta} & ... & b_{\alpha,\beta} \vspace{0.3cm}\\
    		... & ... & ... \vspace{0.3cm}\\
    		b_{\alpha,\beta} & b_{\alpha,\beta} & ... & b_{\alpha,\beta} \vspace{0.3cm}\\
    		\end{bmatrix}		
    	}_{L_{\beta} \textrm{ times} }}
    \right\}
    \,\scriptstyle{L_{\alpha} \textrm{ times}}
\end{align}
\vspace{0.3cm}

The representation of the bipartite networks as a set of nodes $V_{\mathcal{B}}$ and a set of edges $E_{\mathcal{B}}$ can be written as:
\begin{align}
    \left\{
    \begin{array}{ll}
    G_{\mathcal{B}} = (V_{\mathcal{B}}, E_{\mathcal{B}})\\
    V_{\mathcal{B}} = \{v_{k}^{\alpha}, k = 1, ..., n_{\alpha}\} \cup \{v_{l}^{\beta}, l = 1, ...,n_{\beta}\}\\
    E_{\mathcal{B}} = \{e_{k, l}^{\alpha \beta} \; k = 1, ..., n_{\alpha} \; , \; l = 1, ..., n_{\beta} \; ; \; (b_{\alpha, \beta})_{k,l} \neq 0 \} \\
    \end{array}
    \right.
\end{align}

\noindent It is to note that if the bipartite networks are undirected, $b_{\beta, \alpha}^{T} = b_{\alpha, \beta}$ and $B_{\beta, \alpha}^{T} = B_{\alpha, \beta}$. \\

\subsubsection*{Universal multilayer networks unify the representation of heterogeneous multiplex networks}
\hfill\\

We previously defined the Supra-adjacency matrices of each multiplex network and the Bipartite network matrices connecting the different multiplex networks. We now introduce the Supra-heterogeneous adjacency matrix, denoted by $\mathcal{S}$. This matrix, defined in equation [10], collects the $N$ Supra-adjacency matrices representing each multiplex network, $\mathcal{A}_{1}, \mathcal{A}_{2}, ..., \mathcal{A}_{N}$, and the $N*(N-1)$ Bipartite network matrices connecting each multiplex network, $B_{1, 2}, B_{1, 3}, ... , B_{1, N}, B_{2, 1}, ... , B_{N, N-1}$.

\begin{align}
    \mathcal{S} = 
    \begin{bmatrix}
    \mathcal{A}_{1} & B_{1, 2} & $ ... $ & B_{1, N} \vspace{0.3cm} \\
    B_{2, 1} & \mathcal{A}_{2} & $ ... $ & B_{2, N} \vspace{0.3cm} \\
    $ ... $ & $ ... $ & $ ... $ & $ ... $ \vspace{0.3cm} \\
    B_{N, 1} & B_{N, 2} & $ ... $ & \mathcal{A}_{N} \\
    \end{bmatrix}
\end{align}

We can also define the Supra-heterogeneous adjacency matrix as a set of nodes and edges: 

\begin{align}
    \left\{
    \begin{array}{ll}
    G_{\mathcal{S}} = (V_{\mathcal{S}}, E_{\mathcal{S}})\\
    V_{\mathcal{S}} = \bigcup\limits_{k = 1}^{N} \{v^{\alpha_{k}}_{k, i}, i = 1, ..., n_{k}, \alpha_{k} = 1, ..., L_{k}\} \\
    E_{\mathcal{S}} = \bigcup\limits_{k = 1}^{N} (\{e_{i, j}^{\alpha_{k}, \alpha_{k}}, i,j = 1, ..., n_{k}, (A_{k}^{[\alpha_{k}]})_{i, j} \neq 0 \} \\ 
    \hspace{0.5cm} \cup \{e_{i, i}^{\alpha_{k}, \beta_{k}}, i = 1, ..., n_{k}, \alpha_{k} \neq \beta_{k} \; , \; \alpha_{k},\beta_{k} = 1, ..., L_{k} \}) \\
    \hspace{0.5cm} \cup \bigcup\limits_{k,l = 1; k \neq l}^{N} \{e_{i, j}^{\alpha_{k}, \alpha_{l}}, i = 1, ..., n_{k}, j = 1, ..., n_{l}, (B_{k, l})_{i,j} \neq 0\}\\
    \end{array}
    \right.
\end{align}

\subsubsection*{The normalization of the Supra-heterogeneous adjacency matrix ensures the convergence of the RWR to the steady-state}
\hfill\\

The most complex issue is the normalization of the Supra-heterogeneous adjacency matrix into a Transition rate matrix that can be used in equation $[4]$. The normalization allows obtaining a Stochastic matrix that guarantees the convergence of the RWR to the steady-state \cite{Langville2006}(see elements of proof in Supplementary Note 1.A.1). It is important to note that we have chosen a column normalization. The resulting normalized matrix, denoted by $\mathcal{\widehat{S}}$ is defined in equation [12]. We generalized the formalism of Li and Patra \cite{Li2010} established for two heterogeneous monoplex networks (Supplementary Note 1.D). This generalization to universal multilayer networks is done thanks to the intra- and inter- multiplex network normalizations defined in equations [13-14], with $\alpha \in [\![1, N]\!]$, $\beta \in [\![1, N]\!]$. In addition, $c_{i_{\alpha}}$ is the number of bipartite networks in which the node $i_{\alpha}$ appears as source of the multiplex network $\alpha$ denoted by $M_{\alpha}$.

\begin{align}
    \mathcal{\widehat{S}} = 
    \begin{bmatrix}
    \widehat{S}_{11} & \widehat{S}_{12} & $ ... $ & \widehat{S}_{1N} \vspace{0.3cm}\\
    \widehat{S}_{21} & \widehat{S}_{22} & $ ... $ & \widehat{S}_{2N} \vspace{0.3cm}\\
    $ ... $ & $ ... $ & $ ... $ & $ ... $ \vspace{0.3cm}\\
    \widehat{S}_{N2} & \widehat{S}_{N2} & $ ... $ & \widehat{S}_{NN}
    \end{bmatrix}
\end{align}

In equation [13], $\widehat{S}_{\alpha \alpha}$ defines the transition probabilities inside a given multiplex network. In the case of a multiplex network, if a node has no bipartite interactions with nodes from another multiplex networks, we can use the standard normalization. If bipartite interactions exist, then the normalization takes into account the probability that the walker can stay in the multiplex network $(1-\sum_{\beta = 1}^{c_{i_{\alpha}}}\lambda_{\alpha \beta})$. In equation [14], $\widehat{S}_{\alpha \beta}$ defines the transition probability between two different multiplex networks. There are here three possibilities. If the node has no bipartite interactions, the transition probability is equal to zero. If the node has bipartite interactions, the transition probability is equal to the standard normalization weighted by the jump probability $(\lambda_{\alpha \beta})$. Finally, if the node exists only in the bipartite network, the normalization corresponds to the standard normalization weighted by a modified jump probability. This normalization takes into account all the bipartite interactions of the considered node.

\begin{align}
    \widehat{S}_{\alpha \alpha}(i_{\alpha},j_{\alpha}) = \left\{
    \begin{array}{ll}
    \frac{A_{\alpha}(i_{\alpha},j_{\alpha})}{\sum\limits_{k_{\alpha}=1}^{n_{\alpha}}A_{\alpha}(i_{\alpha},k_{\alpha})} & \hspace{-1.2cm} \mbox{if} \; \forall \, \beta: \sum\limits_{k_{\beta}=1}^{n_{\beta}}B_{\alpha, \beta}(i_{\alpha},k_{\beta}) = 0 \\[20pt]
    \frac{(1-\sum\limits_{\beta = 1}^{c_{i_{\alpha}}}\lambda_{\alpha \beta})*A_{\alpha}(i_{\alpha},j_{\alpha})}{\sum\limits_{k_{\alpha}=1}^{n_{\alpha}}A_{\alpha}(i_{\alpha},k_{\alpha})} & \mbox{Otherwise}
    \end{array}
    \right.
\end{align}

\begin{align}
    \widehat{S}_{\alpha \beta}(i_{\alpha},j_{\beta}) = \left\{
    \begin{array}{ll}
    \frac{\lambda_{\alpha \beta} B_{\alpha,\beta}(i_{\alpha},j_{\beta})}{\sum\limits_{k_{\beta}=1}^{n_{\beta}}B_{\alpha,\beta}(i_{\alpha},k_{\beta})} & \hspace{-1.2cm} \mbox{if} \sum\limits_{k_{\beta}=1}^{n_{\beta}}B_{\alpha,\beta}(i_{\alpha},k_{\beta}) \neq 0 \\[20pt]
    \frac{\frac{\lambda_{\alpha \beta}}{\sum\limits_{\beta = 1}^{c}\lambda_{\alpha \beta}} \sum\limits_{i_{\alpha} = 1}^{c} B_{\alpha,\beta}(i_{\alpha},j_{\beta})}{\sum\limits_{i_{\alpha} = 1}^{c} \sum\limits_{k_{\beta}=1}^{n_{\beta}}B_{\alpha,\beta}(i_{\alpha},k_{\beta})} & \mbox{if $i_{\alpha}$ not in $M_{\alpha}$} \\[20pt]
    0 & \hspace{-3cm} \mbox{Otherwise}
    \end{array}
    \right.
\end{align}

The normalization allows including the parameters $\lambda_{\alpha \beta}$ to jump between the multiplex networks (Fig. 2). In other words, these parameters weight the jumps from one multiplex network $\alpha$ to another multiplex network $\beta$, if the bipartite interaction exists. Moreover, the standard probability condition of normalization imposes that $\sum_{\alpha=1}^{N} \lambda_{\alpha \beta} = 1, \forall \, \beta$, where $N$ represents the number of multiplex networks. Finally, the RWR equation on universal multilayer networks is defined as:

\begin{align}
    {\bf p}_{t+1}^{T} = (1-r)\widehat{S}{\bf p}_{t}^{T} + r{\bf p}_{0}^{T}.
\end{align}

\subsubsection*{RWR initial probability distribution in universal multilayer networks}
\hfill\\

The initial probability distribution ${\bf p}_{0}$ from equation $[15]$, which contains the probabilities to restart on the seed(s), can be written in its general form as follows: 

\begin{align}
    {\bf p}_{0}^{T} = 
    \begin{bmatrix}
    \eta_{1} \bar{\bf{v}}^{1}_{0}\\
    \eta_{2} \bar{\bf{v}}^{2}_{0}\\
    ... \\
    \eta_{N} \bar{\bf{v}}^{N}_{0}
    \end{bmatrix}
\end{align}

\noindent where $\eta_{k}$ is the probability to restart in one of the layers of the multiplex network $k$, and $\bar{\bf{v}}^{k}_{0}$ is the initial probability distribution of the multiplex network $k$. The size of $\bar{\bf{v}}^{k}_{0}$ is equal to $(L_{k}*n_{k})$, where $L_{k}$ is the number of layers in the multiplex network $k$ and $n_{k}$ is the number of nodes in the multiplex network $k$. We constraint the parameter $\eta$ with the standard condition of normalization of the probability that imposes $\sum_{k=1}^{N} \eta_{k} = 1$.
We defined another parameter, $\tau$, to take into account the probability of restarting in the different layers of a given multiplex network. This parameter includes $\tau_{kj}$, where $k$ corresponds to the index of the multiplex network, and $j$ to the index of the layer of the multiplex network $k$ (Fig. 2). In other words, $\tau_{kj}$ corresponds to the probability to restart in the $j^{th}$ layer of the multiplex network $k$.
Finally, $\bar{\bf{v}}^{k}_{0}$ is defined as follows: $\bar{\bf{v}}^{k}_{0} = [\tau_{k1}{\bf {v}}^{k}_{0}, \tau_{k2}{\bf {v}}^{k}_{0}, ..., \tau_{kL_{k}}{\bf {v}}^{k}_{0}]^{T}$, with ${\bf {v}}^{k}_{0}$ being a vector with $1/\omega_{k}$ in the position(s) of seed(s) and zeros elsewhere, and $\omega_{k}$ being the number of seeds in the multiplex network $k$. The standard condition of normalization of the probability gives the constraint: $\sum_{j=1}^{L_{k}} \tau_{kj} = 1$, $\forall \, k$.

\subsection*{Numerical implementation: MultiXrank}
\hfill\\

Our RWR on universal multilayer networks is implemented as a Python package called MultiXrank (Supplementary Note 2). MultiXrank has an optimized implementation. Default parameters allow exploring homogeneously the multilayer network (Supplementary Note 1.B). The running time of the package depends on the number of edges of the multilayer network (complexity analyses in Supplementary Note 2.A). \\
The package is available on GitHub \href{https://github.com/anthbapt/MultiXrank}{github/MultiXrank}, and can be installed with standard pip installation command: \href{https://pypi.org/project/MultiXrank}{pypi/MultiXrank}.

\subsection*{Evaluations}
\hfill\\

We evaluated the performances of MultiXrank using two different multilayer networks. The first one is a large biological multilayer network composed of two multiplex networks and one monoplex network. It contains a gene multiplex network gathering gene physical and functional relationships, a drug multiplex network containing drug clinical and chemical relationships, and a disease monoplex network representing disease phenotypic similarities. Each monoplex/multiplex network is connected to the others thanks to bipartite networks containing gene-disease, drug-gene, and drug-disease interactions (Supplementary Note 3.B). The second multilayer network is composed of three multiplex networks. It contains a French airports multiplex network, a British airports multiplex network, and a German airports multiplex network. In each multiplex network, the nodes represent the airports of each country and the edges represent the national flight connections between these airports for three different airline companies. The three multiplex networks are linked with bipartite networks corresponding to transnational flight connections (Supplementary Note 3.A).

We designed a Leave-One-Out Cross-Validation (LOOCV) protocol inspired by F.Mordelet and J.P.Vert \cite{Mordelet2011} and A.Valdeolivas et al. \citep{Valdeolivas2018}. 
In this protocol, we systematically leave-out some known associations and assess the reconstruction of this left-out data using the data remaining in the network (Supplementary Note 4.A and Fig. S9).
In the case of the biological multilayer network,we systematically left-out known gene-disease associations. More specifically, for each disease associated with at least two genes, each gene is remove one-by-one and considered as the left-out gene. The remaining gene(s) associated with the same disease are used as seed(s). When the disease network is considered in the evaluation, the disease node is used as seed together with the gene node(s).
The RWR algorithm is then applied, and all the network nodes are scored according to their proximity to the seed(s). The rank of the gene node that was left-out in the ongoing run is recorded. The perfect ranking for the left-out gene is 1; the closer the rank is to 1, the better the prediction. The gene left-out process is repeated iteratively for all the genes. Finally, the Cumulative Distribution Function (CDF) of the ranks of the left-out genes is plotted (Fig. 3). The CDF displays the ratio of left-out genes that are ranked by the RWR within the top-$K$ ranked gene nodes. The CDFs are used to evaluate and compare the performance of the RWR applied to different combinations of biological networks: the protein-protein interactions (PPI) network alone, the gene multiplex network, the multilayer network composed of the gene multiplex and the disease monoplex networks, and the multilayer network composed of the gene and drug multiplex networks and the disease monoplex network (Fig. 3a). 

We observed that considering multiple sources of network data is always better than considering the PPI alone. In addition, considering multilayer information is better than considering only the gene multiplex network. However, the increased performances in the LOOCV seem to arise only from the gene multiplex network with the disease monoplex network (and associated gene-disease bipartite network). Indeed, the addition of the drug multiplex network (and associated drug-gene and drug-disease bipartite networks) to the multilayer system does not increase the performances (Fig. 3a).

We repeated the same LOOCV protocol for the airports multilayer network, in which the left-out nodes are French airport nodes associated with a given British airport node. Here, the behavior is different, as adding the third multiplex network containing German airports connections (and associated French-German and British-German bipartite networks) increases the performances of the RWR to predict the associations between French and British airports (Fig. 3b).

To better understand these different behaviors, we examined in detail the amount of common nodes (called overlaps) existing between the nodes of the different bipartite networks.
We observed that only 23\% of the genes from the gene-disease bipartite network are present in the drug-gene bipartite network. Similarly, only 5\% of the diseases from the gene-disease bipartite network are present in the disease-drug bipartite network (Fig. S10). 
Given these low overlaps, the drug multiplex network might not contribute significantly to connecting gene and disease nodes during the random walks. This might explain why adding the drug multiplex network does not improve the performances of the LOOCV. Contrarily, the bipartite networks of the airport multilayer network displays high overlaps (Fig. S10). These high overlaps might explain why the addition of the third multiplex network in this case increases the predictive power (Fig. 3b).

To validate the proposed central role of bipartite networks in the RWR performances, we artificially increased the connectivity of the gene-drug and disease-drug bipartite networks before applying the same LOOCV protocol. To this goal, we added artificial transit drug nodes linking existing gene-disease associations (strategy described in Supplementary Note 4C and Fig. S12). We observed that these artificially added transit nodes increased drastically the performances of the LOOCV (Fig. 3c). The same phenomenon is observed for the airports multilayer network (Fig. 3d). In addition, we checked if random perturbations in these artificially enhanced bipartite networks would decrease the performances of the LOOCV. To do so, we progressively randomized the edges in the bipartite networks with artificially increased connectivity, until obtaining completely random bipartite networks. We observed that the progressive randomization of the bipartite networks continuously decreases the predictive power of the RWR up to obtaining the same performances as with only two multiplex networks (Fig. S13.A for the airport multilayer networks and S13.B for the biological multilayer networks).

Finally, we repeated all these evaluations using a standard Link Prediction (LP) protocol (Supplementary Note 4.B). LP has already been used to measure the predictive power of RWR methods \cite{Zhou2020}. In the LP protocol, we systematically removed gene-disease edges from the gene-disease bipartite network, and predicted the rank of the removed gene using the disease as seed in the RWR. The LP protocol is applied on the airport multilayer network by removing a French-British edge from the French-British bipartite network, and predicting the rank of the French airport using the British airport node as a seed in the RWR. We overall observed similar behaviors as in the LOOCV (Fig. S11 and S14).

Importantly, the LOOCV and LP protocols can be used to evaluate the pertinence of adding new multiplex networks in a multilayer network or new network layers in a multiplex network. Both evaluation protocols are available within the MultiXrank package.

\subsection*{Parameter space exploration}
\hfill\\

We next evaluated the stability of MultiXrank output scores upon variations of the input parameters. We illustrate this exploration of the parameter space with the biological multilayer network composed of the gene multiplex network and the disease monoplex network.
We first compared the top-5 and top-100 gene and disease nodes prioritized by MultiXrank using 125 different sets of parameters (see Supplementary Note 5 for the definition of the sets of parameters). We observed that the top-ranked gene nodes vary more depending on the input parameters than the top-ranked disease nodes (Fig. 4a).

To better understand the stability of the output scores upon variations of the input parameters, we proposed a protocol based on 5 successive steps: i) definition of the sets of parameters, ii) construction of a matrix containing the similarities of the RWR output scores obtained with each set of input parameters, using a the similarity measure defined in equation [17]. The similarities are computed for each type of node independently (i.e., for gene and disease nodes independently). 

\begin{equation}
	\Theta_{\gamma \sigma}^{k} = \sum_{j = 1}^{n_{k}} \frac{\sqrt{(\frac{1}{[({\bf r}_{\gamma}^{k})_{j} - ({\bf r}_{\gamma \sigma}^{k})_{j}]})^{2} + (\frac{1}{[({\bf r}_{\sigma}^{k})_{j} - ({\bf r}_{ \sigma \gamma}^{k})_{j}]})^{2}} }{ (\frac{({\bf r}_{\gamma}^{k})_{j} + ({\bf r}_{\sigma}^{k})_{j}}{2})^{2}}
\end{equation}

\noindent where $\gamma$ and $\sigma$ define two sets of parameters, $n_{k}$ is the number of nodes associated with the multiplex network $k$. In addition, ${\bf r}_{\gamma}^{k}$ (resp. ${\bf r}_{\sigma}^{k}$) is the rank output scores distribution that associates with each node its rank given by the RWR with the set of parameters $\gamma$ (resp. $\sigma$) for the multiplex network $k$. Finally, ${\bf r}_{\gamma \sigma}^{k}$ (resp. ${\bf r}_{\sigma \gamma}^{k}$) gives to each node of the output scores distribution obtained by the set of parameters $\gamma$ (resp. $\sigma$) (in the multiplex network $k$) their rank in the distribution $\sigma$ (resp. $\gamma$). 

We next computed a consensus Similarity matrix with a normalized euclidean norm of each individual Similarity matrix (equation [18]).

\begin{equation}
	\Theta_{\gamma \sigma} = \sqrt{\sum_{k = 1}^{N} \frac{(\Theta_{\gamma \sigma}^{k})^2}{n_{k}}}
\end{equation}

\noindent where $N$ is the number of multiplex networks.

The next step is iii) projection of the consensus Similarity matrix into a Principal Component Analysis (PCA) space (Fig. 4b). In this PCA space, each dot represents the output scores resulting from a set of parameters. Then, iv) clustering (using k-means on the two first principal components) to identify sub-regions containing similar RWR output scores. Finally, v) comparing the top-ranked nodes obtained with the set of parameters belonging to each cluster (Fig 4c., Supplementary Note 5).

We applied this protocol to evaluate the output scores obtained by MultiXrank on the previously defined biological multilayer network composed of the gene multiplex network and the disease monoplex network, using 125 different combinations of parameters (Fig. 4, supplementary Fig. S16). We projected the consensus Similarity matrix into a PCA space and identified 8 clusters (Fig. 4b). To illustrate the behavior inside clusters, we concentrated our analyses on the two clusters defined in the bottom left subspace (clusters number 4 and 6, zoom-in Fig. 4b). The top-100 ranked gene and disease nodes inside each of the two clusters are overall similar (Fig. 4c). This means that, even if the node prioritization can be sensitive to input parameters, we can identify regions of stability in the parameter space. Moreover, the protocol allows identifying the monoplex/multiplex networks that generate most variability in the output scores upon changes in the input parameters. 

We applied the parameter space exploration protocol to other multilayer networks and observed diverse behaviors, from highly variable top-rankings and scattered projections in the PCA space for the airport multilayer network (supplementary Fig. S15) to robust top-rankings with well-clustered projections in the PCA space for the biological multilayer network composed of 3 types of nodes (genes, diseases and drugs, supplementary Fig. S16). Overall, our parameter space study reveals different sensitivities to input parameters depending on the multilayer network explored. The protocol is available within the MultiXrank package and can be used to characterize in-depth the sensitivity to input parameters of any multilayer network.

\section*{Discussion}
\hfill

Multilayer networks are nowadays very popular, in particular because they allow capturing a larger part of real and engineered systems. In biology, multilayer networks integrating multiscale sources of heterogeneous interactions provide a more comprehensive picture of biological system functionalities. However, data representation as multilayer networks must be accompanied by the development of tools allowing their exploration. Many efforts are thereby dedicated to extend classical network theory algorithms to multilayer systems \cite{Kivelae2014, Boccaletti2014}. These algorithms include for instance clustering algorithms \cite{Huang2021}, Graph Convolutional Networks \cite{Ghorbani2019, Shanthamallu2020} or meta-path based methods \cite{Zhang2019, Himmelstein2017}. Other important network exploration algorithms, such as diffusion kernels or methods based on random walk, are based on the principle of network propagation \cite{Boccaletti2014}. The methods based on random walk, such as PageRank, biased random walk or Random Walk with Restart (RWR), are widely used in network science. They are indeed versatile: the random walk output scores can be used directly for node prioritization and subnetwork extraction, but can also be used as input for downstream analyses, for instance for supervised classification or node embedding \cite{PioLopez2021}.

Different random walk methods have been adapted to consider multilayer networks.
However, a large variety of multilayer networks exist, from multiplex to temporal networks, for instance. To the best of our knowledge, network exploration algorithms that have been adapted to handle multilayer networks can usually be applied only to specific categories of multilayer networks, such as multiplex networks composed of the same set of nodes.

We present here MultiXrank, a tool that proposes an optimized and general formalism for RWR on universal multilayer networks. MultiXrank can be applied to explore multilayer networks composed of any combination of multiplex, monoplex or bipartite networks, and all the network edges can be directed and/or weighted. To the best of our knowledge, any type of multilayer networks could be represented with our formalism, even if it might sometimes require some adaptations. We illustrated the use of MultiXrank with RWR on biological and airport multilayer networks and thereby provide guidelines for users. 
Even if one's initial intuition in data analysis could be that "more data is better", the addition of interaction network layers also brings additional degrees of freedom \cite{Kivelae2014}. To evaluate the pertinence of the addition of multiplex networks or the addition of layers in a multilayer system, MultiXrank includes a systematic evaluation protocol based on Leave-One-Out-Cross-Validation and Link Prediction. Overall, our results show that adding networks data does not always increase the predictive power of the RWR, as already suggested by previous studies \cite{Choobdar2019}. Our evaluation protocol can be used, for the first time to our knowledge, to evaluate in-depth the signal-to-noise of multilayer system combinations. 
Finally, we complemented MultiXrank with a parameter space exploration protocol to measure the influence of varying the input parameters on the global stability of the output scores. It is to note that this parameter space exploration protocol is universal and can be used to study any complex system exploration approach providing scores as outputs.

The output scores of MultiXrank can be used in a wide variety of downstream analyses. For instance, shallow embedding methods need similarity measures for the optimization of the loss function \cite{Hamilton2018, PioLopez2021}. MultiXrank can produce such a similarity measure respecting the global topology of the multilayer network. An interesting application could be to use MultiXrank output scores for embedding and evaluate the predictive power of the gene-disease association prediction task. Indeed, the embedding is expected to be more robust to the noise than the direct network space\cite{Nelson2019}.

The MultiXrank package can be applied to any kind of multilayer network such as social, economic, or ecological multilayer networks. MultiXrank is optimized and can handle multilayer networks containing up to millions edges. To consider billion-scale network problems, several strategies could be considered, such as the Block Elimination Approach for RWR (BEAR) that can be exact or approximate \cite{Shin2015} or the Best of Preprocessing and Iterative approaches (BEPI) that is an approximate approach \cite{Jung2017}.

\section*{Data availability}

All the data and the code used in the article are available on an OSF repository: \href{https://osf.io/zsmua/}{https://osf.io/zsmua} (DOI 10.17605/OSF.IO/ZSMUA). This repository includes all the results obtained in the article.

\section*{Code availability}

The package is available on GitHub \href{https://github.com/anthbapt/MultiXrank}{github/MultiXrank}, can be installed with standard pip installation command: \href{https://pypi.org/project/MultiXrank}{pypi/MultiXrank}, and is associated with complete documentation: \href{https://multixrank-doc.readthedocs.io/en/latest}{https://multixrank-doc.readthedocs.io/en/latest}.

\section*{References}
\bibliography{multixrank_biblio}

\section*{Acknowledgments}
\footnotesize{
The project leading to this preprint has received funding from the « Investissements d'Avenir » French Government program managed by the French National Research Agency (ANR-16-CONV-0001), from Excellence Initiative of Aix-Marseille University - A*MIDEX and from the Inserm Cross-Cutting Project GOLD.}

\section*{Author contributions}
\footnotesize{
A.Bap. and A.Bau. designed research; A.Bap. performed research; A.Bap. analyzed data; A.Bap. and A.G contributed to packaged code; A.Bap. and A.Bau. wrote the paper.}

\section*{Competing interests}
\footnotesize{
The authors declare no competing interests.}

\newpage

\begin{figure*}[ht!]
	\begin{center}
		\captionsetup{justification=centering}
		\centering
		\includegraphics[width=8.5cm,height=8.5cm,keepaspectratio]{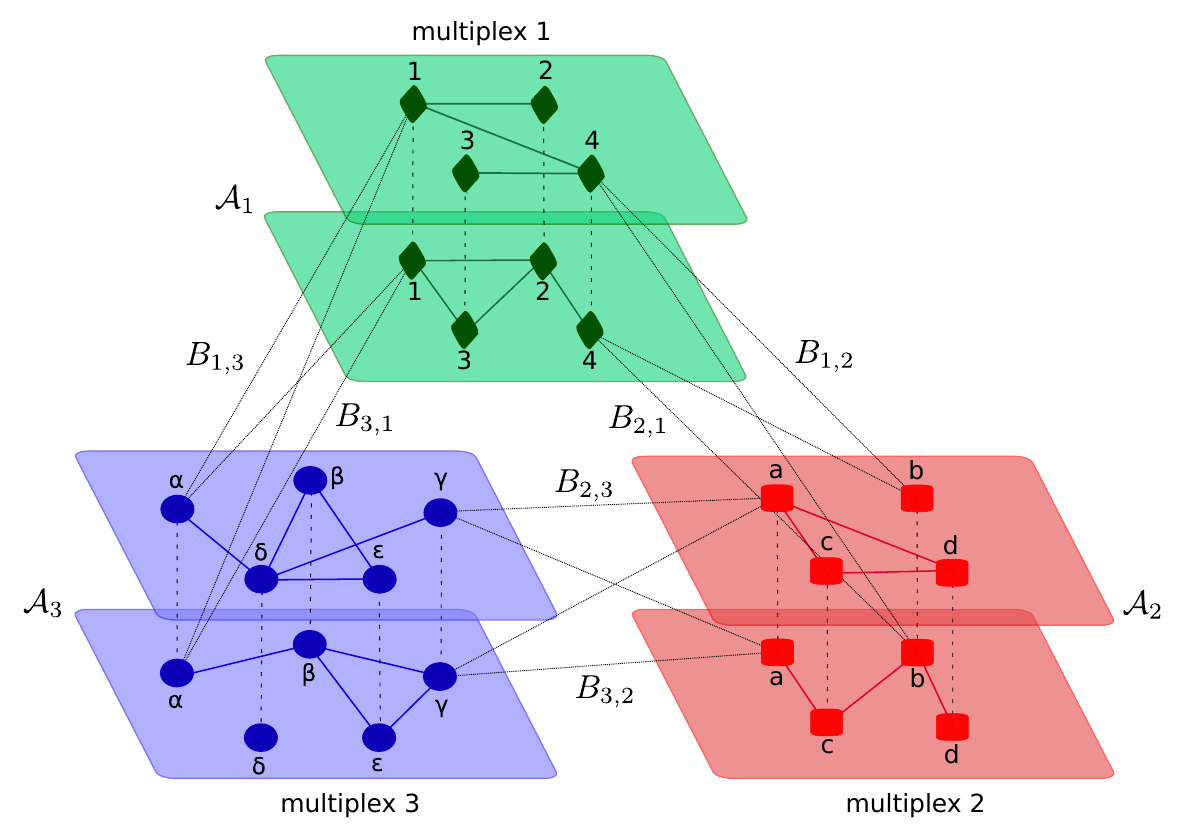}
		\caption{A universal multilayer network. A universal multilayer network composed of three multiplex networks (green, blue and red multiplex networks). Each multiplex network contains different types of nodes (denoted 1 to 4, $\alpha$ to $\epsilon$, and a to d, respectively). Their corresponding Supra-adjacency matrices are denoted by $\mathcal{A}_{i}$. The three multiplex networks are linked by six bipartite networks (represented here as bipartite interactions for the sake of visualization). The corresponding Bipartite network matrices are denoted by $B_{i,j}$. It is to note that a connection between a node $i$ in the multiplex network $\alpha$ and node $j$ in multiplex network $\beta$ imposes the creation of edges between all replicas of node $i$ present in the different layers of the multiplex network $\alpha$ and all replicas of node $j$ present in the different layers of multiplex network $\beta$. All the edges of the universal multilayer networks can be weighted and/or directed.}
	\end{center}
\end{figure*}

\newpage

\begin{figure*}[ht]
	\captionsetup{justification=centering}
	\centering
	\includegraphics[width=15cm,height=15cm,keepaspectratio]{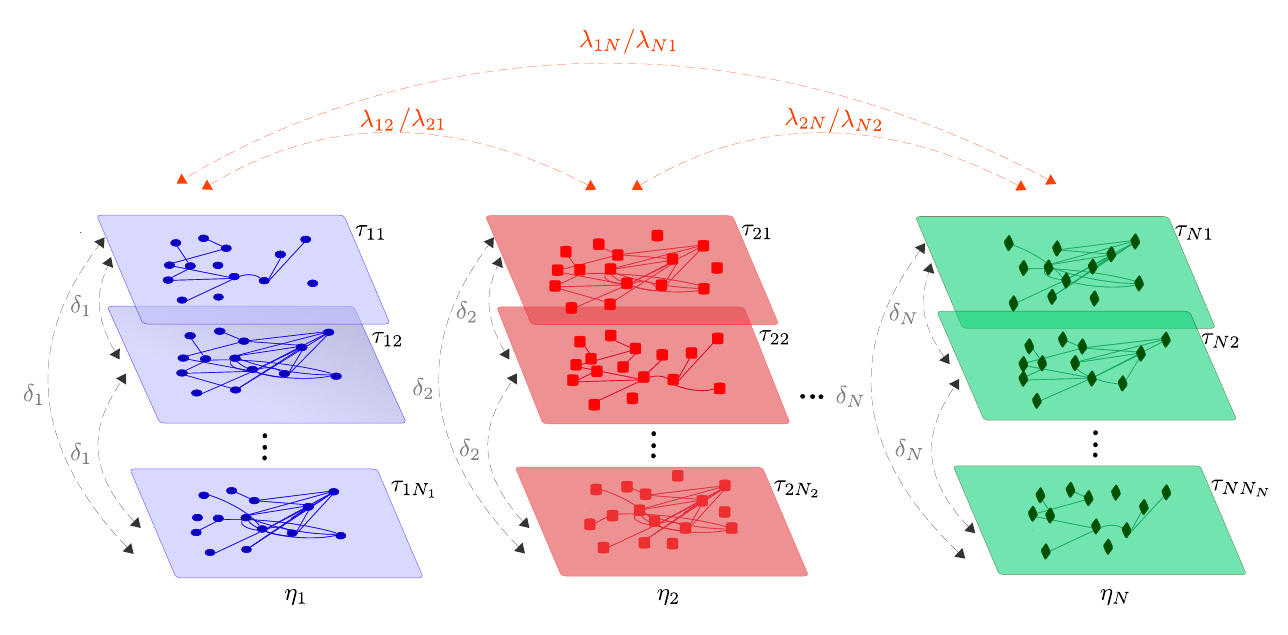}\\
	\caption{MultiXrank Random Walk with Restart parameters. Parameters of the Random Walk with Restart allowing to explore universal multilayer networks composed of $N$ multiplex networks (each composed of several layers containing the same set of (replica) nodes but different edges). The parameters $\delta$ are associated with the probability to jump from one layer to another in a given multiplex network, $\lambda$ with the probability to jump from one multiplex network to another multiplex network, $\tau$ with the probability to restart in a given layer of a given multiplex network, and $\eta$ with the probability to restart in a given multiplex network.}
\end{figure*}

\newpage

\begin{figure*}[ht!]
	\captionsetup{justification=centering}
	\centering
	\includegraphics[width=15cm,height=15cm,keepaspectratio]{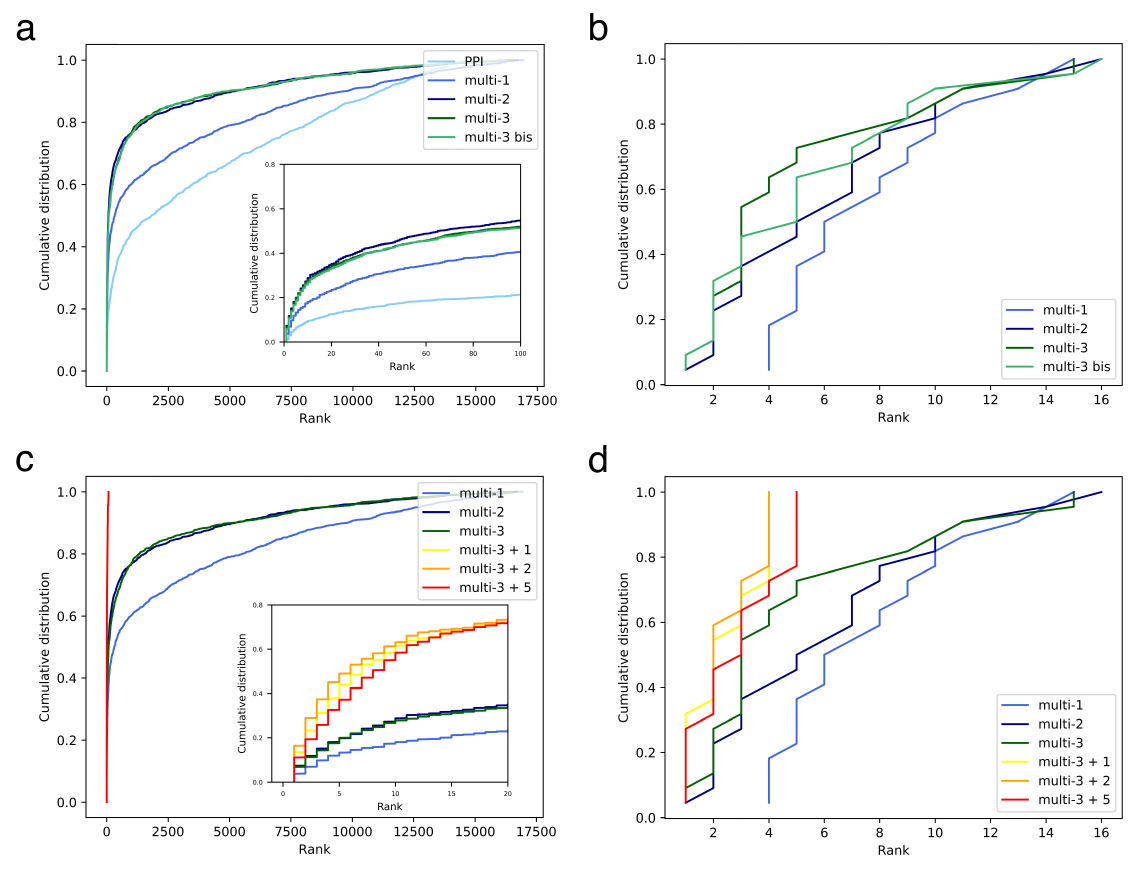}
	\caption{Evaluation and comparison of MultiXrank performances on different combinations of multilayer networks.\\
	a and b: Cumulative Distribution Functions (CDFs) representing the ranks of the left-out nodes in the Leave-One-Out Cross-Validation (LOOCV) protocol. a: focus on different combinations of biological networks: protein-protein interactions network alone (PPI), gene multiplex network (multi-1), multilayer network composed of the gene multiplex network and the disease monoplex network (multi-2), and multilayer network composed of the gene and drug multiplex networks and the disease monoplex network, for two different sets of parameters (multi-3, multi-3 bis). The multilayer networks are connected by the bipartite networks described in the Evaluations section. b: focus on different combinations of airports networks: French multiplex network (multi-1), multilayer network composed of the French and British airports multiplex networks (multi-2), and multilayer network composed of the French, British, and German airports multiplex networks, for two different sets of parameters (multi-3, multi-3 bis). These multilayer networks are connected by the bipartite networks described in the Evaluations section.\\
	c and d: CDFs representing the ranks of the left-out nodes in the LOOCV protocol for the multi-3 multilayer networks described previously with artificially increased connectivity in the gene-drug and disease-drug bipartite networks. c: The connectivity is artificially increased thanks to the addition of 1 (multi3+1), 2 (multi3+2) or 5 (multi3+5) transit drug nodes for each gene-disease association. d: In the airport multilayer network, the connectivity is artificially increased in the French-German and British-German bipartite networks thanks to the addition of 1 (multi3+1), 2 (multi3+2) or 5 (multi3+5) transit German nodes for each French-British airports association. The parameters of the Random Walk with Restart (RWR) are detailed in Supplementary Tables S5-S6.}
\end{figure*}

\newpage

\begin{figure*}[ht]
	\centering
	\captionsetup{justification=centering}
	\includegraphics[width=17.5cm,height=17.5cm,keepaspectratio]{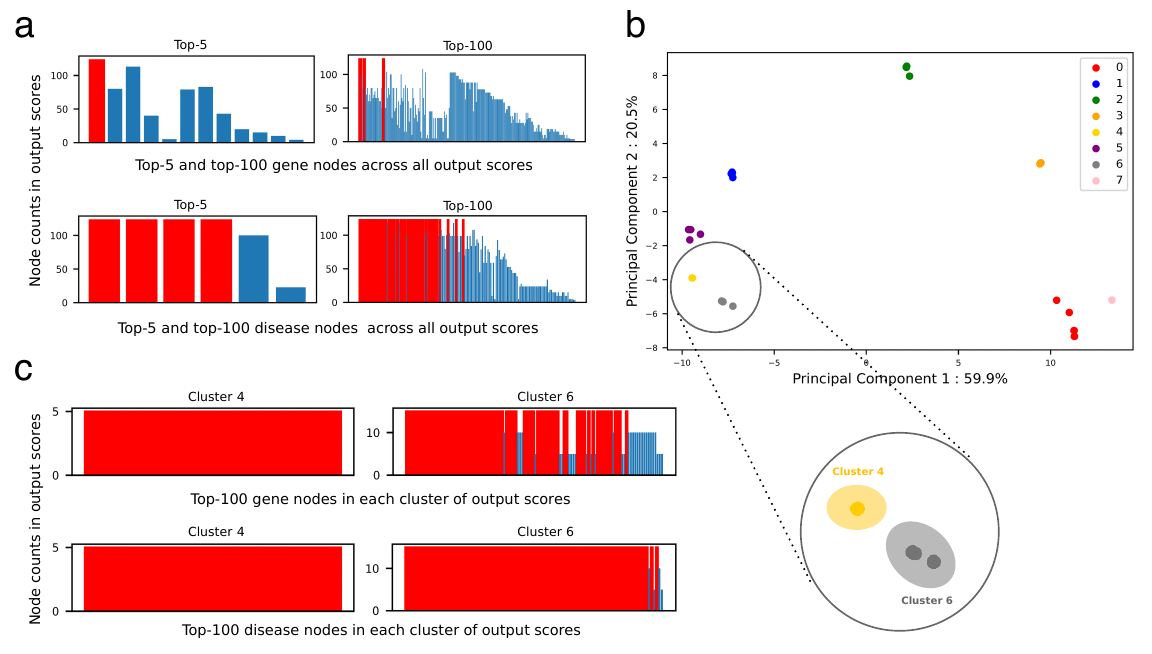}\\
	\caption{Exploration of MultiXrank parameter space. a: Comparison of the top-5 and top-100 nodes ranked by MultiXrank using a biological multilayer networks composed of the gene multiplex network and the disease monoplex network for 125 different sets of parameters. The top-5 or top-100 ranked nodes for each set of parameters are merged, and the number of occurrences of each node are counted. The nodes are represented in bars colored in red when the node is found in all top-5 or top-100 scores, and in blue otherwise. b: Clustering in the Principal Component Analysis (PCA) space of the output scores obtained with MultiXrank on the biological multilayer network composed of the gene multiplex network and the disease monoplex network using 125 different sets of parameters. The zoom-in emphasizes the clusters number 4 and 6. c: Comparison of the top-100 nodes retrieved for the sets of parameters belonging to clusters 4 and 6 defined in b. The bar is colored in red when a node is found in all top-100 scores, and in blue otherwise. The parameters of the Random Walk with Restart (RWR) are detailed in Supplementary Table S7.}
\end{figure*}

\end{document}